\documentclass{article}

\usepackage{graphicx} %
\usepackage[most]{tcolorbox} %
\usepackage[main]{neurips}
\usepackage[utf8]{inputenc} %
\usepackage[T1]{fontenc}    %
\usepackage[colorlinks]{hyperref} 
\usepackage{listings}
\hypersetup{
 linkcolor=[RGB]{58, 116, 7},
 citecolor=[RGB]{58, 116, 7}
}
\usepackage{url}            %
\usepackage{booktabs}       %
\usepackage{amsfonts}       %
\usepackage{nicefrac}       %
\usepackage{microtype}      %
\usepackage{xcolor}         %
\usepackage{subcaption}
\usepackage{color}
\usepackage{transparent}
\usepackage{std_definitions}

\definecolor{succ}{RGB}{218,241,222}
\definecolor{abstract}{RGB}{208,154,165}
\definecolor{mediumseagreen}{RGB}{84,180,73}
\newcommand{\llm}{\mathbb{P}}

\begin{document}

\includegraphics[scale=0.15]{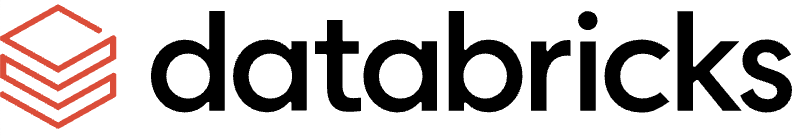}

\begin{tcolorbox}[
width=\linewidth,
colback=abstract!5!white, 
colframe=red!75!black, 
title=
]

\begin{center}
{\Large \textbf{A State-of-the-Art SQL Reasoning Model using RLVR}\vspace*{0.4cm}}

\textbf{The Databricks Mosaic Research and Genie Teams}\footnote{A complete list of contributors is given in the appendix.}\\
\end{center}

Developing custom reasoning models via Reinforcement Learning (RL) that can incorporate organization-specific knowledge has great potential to address problems faced by enterprise customers. In many of these problems, the reward function is verifiable, a setting termed RL with Verifiable Rewards (RLVR). We apply RLVR to a popular data science benchmark called BIRD that measures the ability of an AI agent to convert natural language query for a database to SQL executions.  
We apply a simple and general-purpose training recipe involving careful prompt and model selection, a warm-up stage using our offline RL approach called TAO, followed by rigorous online RLVR training. With no additional training data beyond the BIRD training set and no use of proprietary models, our very first submission to the BIRD leaderboard reached state-of-the-art accuracy on the private test set: 73.56\% without self-consistency and 75.68\% with self-consistency. In the latter case, our model also required fewer generations than the second-best approach. While BIRD is only a proxy task, the simplicity of our framework makes it broadly applicable to enterprise domains such as business intelligence, data science, and coding.\looseness=-1

\end{tcolorbox}

\section{Introduction}

LLMs have become a fixture in a range of enterprise problems including software engineering~\cite{yang2024swe}, data science~\cite{hong2024data}, and more recently even research itself~\cite{oai2025deepresearch}. However, despite sometime impressive performance of off-the-shelf LLMs, they face limitations in doing more bespoke enterprise tasks displaying limitations such as unable to understand organization-specific terminology or use organization-specific concepts, tools, and APIs. Post-training LLMs using approaches such as Reinforcement Learning with Verifiable Rewards (RLVR) ~\cite{allen2024tulu3,hu2025open,wen2025rlvr} provides one way to address this. In this report, we show how we applied RLVR to fine-tune an LLM to achieve state-of-the-art reasoning model on the popular data science benchmark BIRD~\cite{li2023can}.

\begin{figure}
    \centering
    \includegraphics[width=\linewidth]{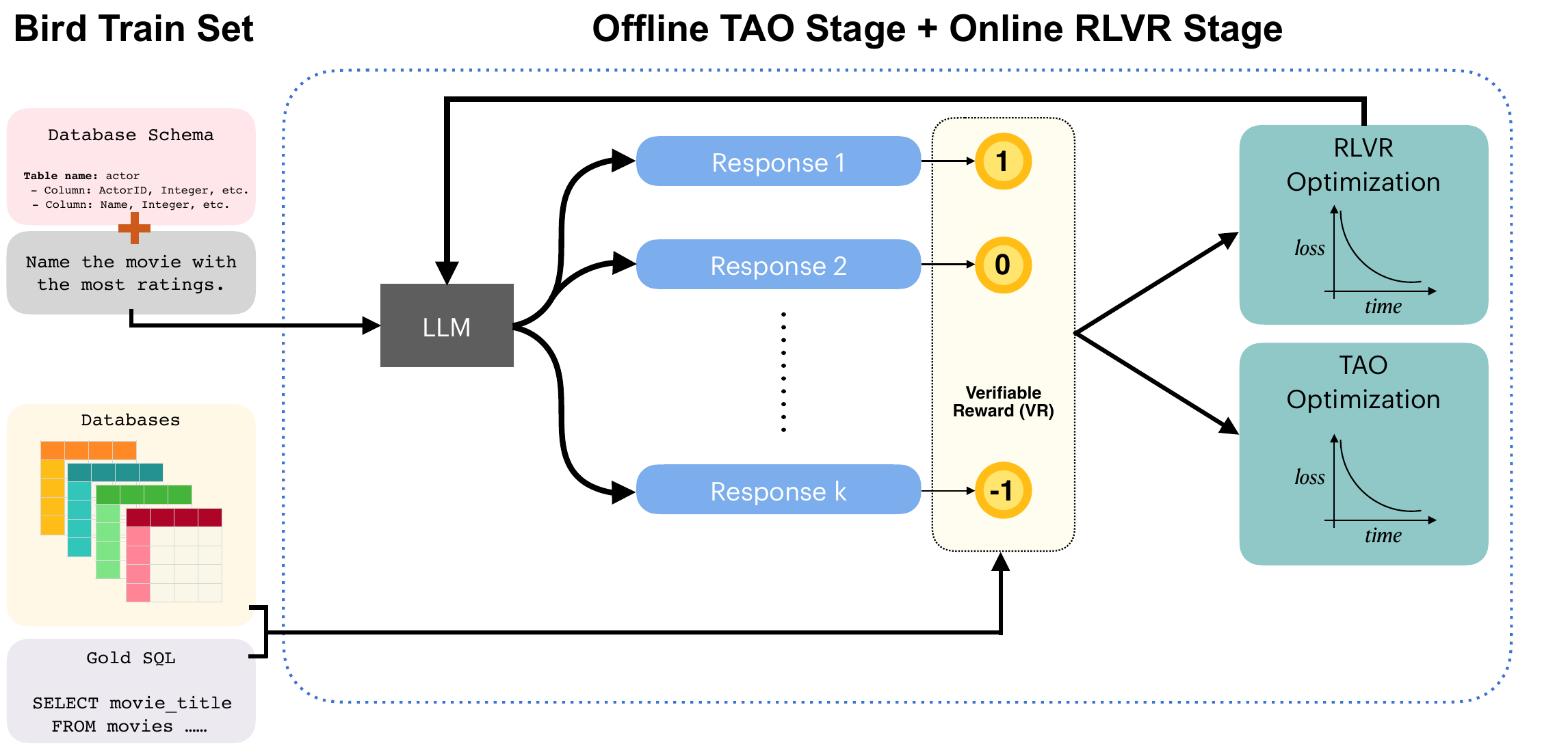}
    \caption{Illustration of our RLVR-based fine-tuning pipeline applied to text-to-SQL. Given a user query (e.g., ``\emph{Name the movie with the most ratings}'') together with the structure of relevant database tables and associated domain knowledge, the goal is to produce SQL that matches the reference (``gold'') SQL shown in the lower left. Our pipeline generates verifiable rewards by sampling $k$ candidate SQL queries from the current model, executing them, and comparing results against the gold SQL. These RLVR rewards are then used to update the model with either TAO our offline RL approach, or an online RLVR optimization.}%
    \label{fig:intro-fig}
\end{figure}

\pref{fig:intro-fig} shows an example of the text2sql task which is a common data science task of allowing a user to query a database. In tasks such as these, off-the-shelf LLMs can struggle for various reasons. For example, different organizations or industries might use specific jargon that may not be present in LLM's knowledge. For example, given a user query \emph{``show me all churned users for 2024?"}, the model may not know how the organization defines \emph{``churned users"}. Or, the model may not be aware of certain preferences such as always sorting certain queries with a given column. And yet other reasons, can be not knowing how to use certain APIs, tools, and know where to find data. Unlike math problems where the training data might be more available on the internet, organizational knowledge might be more protected and less represented on the internet posing challenges for LLMs whose training data may not cover this.

There are several ways to address this knowledge gap. Post-training LLMs through approaches such as RLVR provide one way to address this knowledge gap. In particular, for data science and coding tasks, one can often compute a \emph{verifiable reward} which does not need to be trained and serves as an objective quality metric. For the task in~\pref{fig:intro-fig}, we can measure if a given AI agent's SQL response is correct by matching it against the ground truth SQL response for that user query. Other approaches such as GEPA~\cite{agrawal2025gepa} approach this by optimizing prompts instead of LLM weights. At Databricks we support both these approaches along with a suite of other methods. The preference for a given approach can depend on the type of dataset and other resource requirements. In this report, we focus on RLVR and demonstrate its power by applying it to the BIRD benchmark (\pref{fig:intro-fig}).

Our approach starts by carefully selecting the model, context, and prompt. The prompt can be chosen using prompt optimization approaches such as GEPA~\cite{khattab2023dspy,agrawal2025gepa}. We then perform a two-stage fine-tuning of the model with the chosen context and prompt. First, we warm start the model using Test-time Adaptive Optimization (TAO), an offline RL approach~\cite{tao}. This is followed by fine-tuning the model using Databricks's RLVR service. Our RLVR service supports popular learning approaches such as GRPO~\cite{shao2024deepseekmath}, as well as more powerful alternatives. Finally, we perform self-consistency on top of the fine-tuned model at inference time. This general recipe allowed us to establish a new state-of-the-art accuracy of \textbf{75.68\%} on the private test set of the BIRD benchmark in single-model category with our first submission.\footnote{BIRD benchmark is available here: https://bird-bench.github.io/. All numbers are current as of September 15th, 2025.} Our RLVR-trained model already achieves state-of-the-art in single-model single-LLM call category, whereas self-consistency adds complimentary benefits and gives us the overall best single-model performance. Further, our model generalizes better than other top-submissions and achieves state-of-the-art using fewer self-consistency responses than the second best result.

\section{Overview}

We first provide an overview of our setup and approach.

\paragraph{Reinforcement Learning from Verifiable Rewards (RLVR).}  RLVR is a post-training paradigm of applying reinforcement learning to fine-tune a pre-trained LLM on a given task where the reward function measures an objective truth without needing any parametrized reward models~\cite{allen2024tulu3,hu2025open,wen2025rlvr,rastogi2025magistral}. RLVR is particulary suited to tasks where the LLM is trying to predict facts, mathematical statements, or generating executable code, as its responses can be measured objectively given access to ground truth. This contrasts with non-verifiable tasks that typically measure subjective tasks such as generating poetry, or an essay on a given topic~\cite{christiano2017deep}. Significant focus has been on applying RLVR to maths and coding tasks~\cite{rastogi2025magistral}. An important current challenge is applying RLVR to complex enterprise tasks such as business analysis and data science workflows.

\paragraph{Text2SQL.} In this report, we focus on evaluation on an important data science problem called text2sql~\cite{finegan2018improving,misra2018policy,hwang2019comprehensive,li2023can}. In this task, a user asks a natural language query $x \in \Xcal$ to access a database $d \in \Dcal$. An agent then generates a SQL code $y \in \Ycal$ which is executed on the database and its output is returned to the user. The goal of the agent is to generate a SQL code that best captures the intent of the user query. The main challenge in text2sql is that it requires understanding the query in the context of the database which can be quite complex. For example, the database may have many tables with many columns, column names can be ambiguous, correct column values may need to be found and tables may need to be joined to answer the user query. 

We approach this problem relying on a reasoning-based LLM~\cite{oai2025o3o4mini,shao2024deepseekmath,yang2025qwen3} $\llm$ by encoding the context $x$ and database $d$ in a prompt and then querying the LLM $z \sim \llm(\cdot \mid x, d)$ to generate a reasoning trace $z \in \Zcal$ including the SQL response. The SQL response is marked using delimiters which allows us to extract it from the trace. We focus on how to fine-tune an LLM on this task using RLVR.

We evaluate on the BIRD benchmark which is a popular text2sql benchmark~\cite{li2023can}. A BIRD datapoint includes a user query, an evidence field providing additional instruction, a database containing multiple tables, and a gold SQL code denoting ground truth. We combine user query and evidence into a single unified user query. The evaluation metric for BIRD is a strict 0-1 metric where given the generated SQL code, we execute both the gold SQL and the generated SQL code on the database and check if their outputs match ($1$) or not ($0$). The BIRD dataset consists of a train set of 9,428 examples, a dev set of 1,534 examples, and a test set of 1,789 examples. The test set is not publicly available allowing evaluation without dataset contamination concerns and prevents benchmark hacking~\cite{zhou2023don}. We train the model using the train set and use the dev set performance for model and hyperparameter selection. We do not use any additional training data.

\paragraph{Test-time Computation.} Test-time computation is any approach that spends more inference compute at test time to improve a model's performance. One common test-time computation approach is self-consistency~\cite{wang2022self} which involves generating multiple responses from an LLM and then using a procedure to select the most consistent response. We will use self-consistency on top of our RLVR-ed model.\looseness=-1

\section{Methodology}

\begin{figure}
    \includegraphics[width=\linewidth]{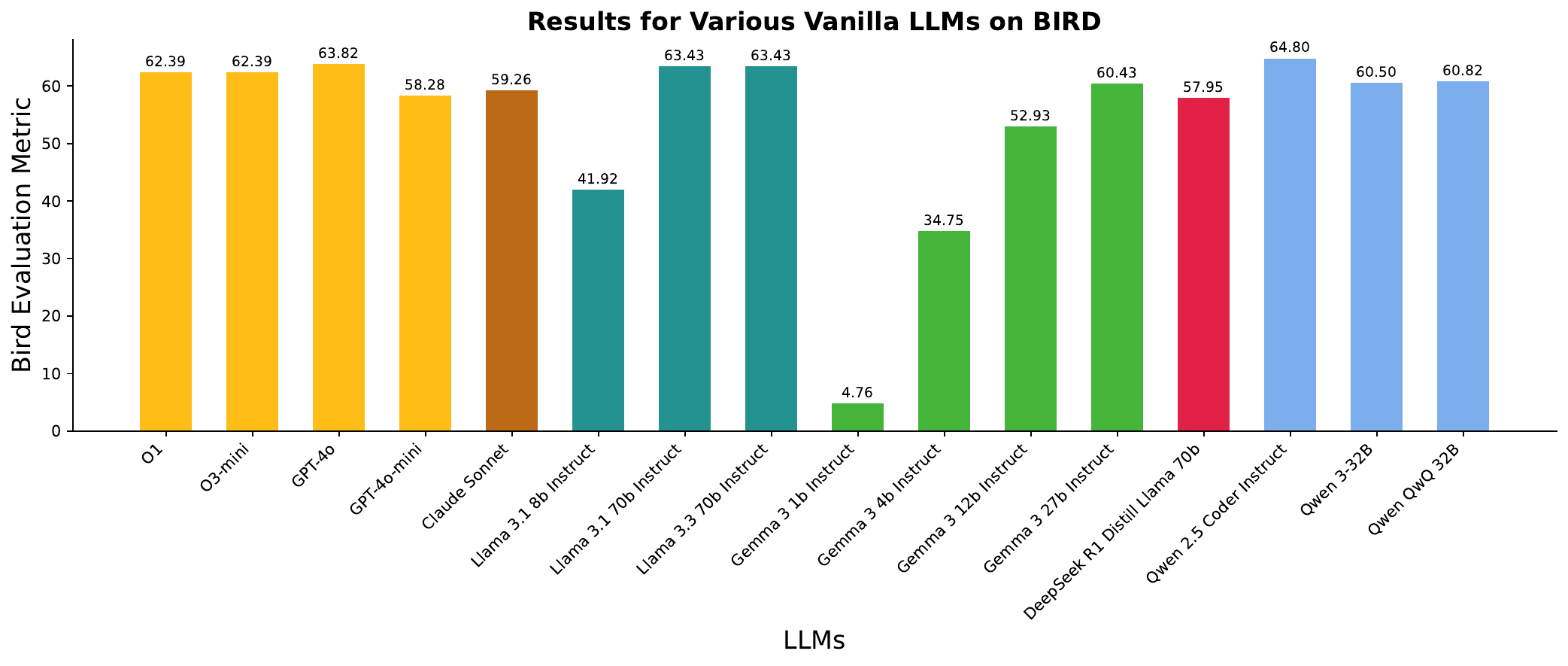}
     \caption{Results on the BIRD dev set across a range of models. We perform greedy decoding for all models except O1 and O3-mini which only permit temperature based decoding.}
    \label{fig:base-results}
\end{figure}

In this section, we provide an overview of our methodology.

\paragraph{Prompt and Model Selection.}

We evaluated a range of open-source models along with closed-source models on a range of different prompts on the BIRD dev set. We found that the Qwen 2.5 32B Coder Instruct model produced the best dev performance. This matches the best result reported in the literature by the OmniSQL study~\cite{li2025omnisql}. We show results in~\pref{fig:base-results}.

Perhaps surprisingly, we found that proprietary LLMs such as GPT-4o, O3, Claude Sonnet do not do as well as Qwen and Llama on this task. This challenges the common belief that proprietary LLMs are always a better choice. We leave further investigation of this for future work.

We found that OmniSQL prompt was a good starting point. However, we made  important modifications to it. Firstly, we changed the database encoding to enable the LLM to understand the dataset more easily. Secondly, we added additional instructions to elicit reasoning since in our early study we found that the model was often not generating any reasoning trace with the  OmniSQL prompt.%

\paragraph{Verifiable Reward.} We define a simple reward for use in both TAO and RLVR training. We take the 0-1 BIRD execution metric and add a penalty of -1 if the model generates a syntactically incorrect SQL code. This simple metric captures the desired evaluation metric up to an added penalty. Our preliminary investigation using reward shaping did not yield any significant improvements.

\paragraph{TAO Training.} We first perform TAO training which is an offline RL approach pioneered at Databricks~\cite{tao}. The goal is to warm start the model for RLVR training by providing it with a good inductive bias. In a single iteration of TAO, we first generate several responses using the current model for every BIRD training datapoint. We then compute a reward for each response using the above verifiable reward. Finally, we perform TAO optimization which performs offline RL on the collected dataset. This entire procedure can be repeated multiple times.

\paragraph{RLVR Training.} We finally perform online RL with verifiable rewards (RLVR) on top of the best TAO-ed model using our Databricks fine-tuning service. We swept through a variety of hyperparameters and noticed that in general the use of lower learning rate and a lower KL-divergence coefficient was desirable. In fact, completely removing the KL divergence term in RLVR did not significantly hurt the performance echoing recent findings in the literature~\cite{rastogi2025magistral}. Our RLVR fine-tuning service performs several improvements on top of known state-of-the-art RLVR approaches such as GRPO~\cite{shao2024deepseekmath}. These improvements address various efficiency concerns such as ensuring the training focuses on challenging problems and to avoid reward saturation. %

\section{Empirical Results and Discussion}
\label{sec:results}

We first present results without test-time compute and then with it.

\subsection{Results without test-time compute}
\pref{tab:bird-results-single-call} shows the top models on the BIRD leaderboard that do not use self-consistency.\footnote{We directly use results from the leaderboard and all leaderboard results are current as of September 15th, 2025.} %
For clarity, in all the tables below we refer to our full model---Qwen2.5-32B-Coder-Instruct fine-tuned with TAO and RLVR---as ``Databricks-RLVR-32B''. We also report results with the base model (``Qwen2.5-32B'') and a TAO-only variant (``Qwen2.5-32B + TAO'') for comparison purposes.

\begin{table}[!htb]
    \centering
    \renewcommand{\arraystretch}{1.3}
    \setlength{\tabcolsep}{10pt}
    \begin{tabular}{l|cc}
        & \multicolumn{2}{c}{\textbf{Accuracy (exact-match)}} \\
        \textbf{Model} & \textbf{Dev set} & \textbf{Test set} \\
        \hline\hline
        \textbf{Databricks-RLVR-32B} & \textbf{70.80} & \textbf{73.56} \\
        Qwen2.5-32B-Coder-Instruct + TAO & 67.40 & -- \\
        Qwen2.5-32B-Coder-Instruct & 64.80 & -- \\
        \hline
        Arctic-Text2SQL-R1-32B~\cite{yao2025arctic} & 70.47 & 71.83 \\
        Arctic-ExCoT-70B~\cite{zhai-etal-2025-optimizing} & 68.51 & 68.53 \\
        Command A~\cite{cohere2025command} & 63.49 & 65.68 \\
        OneSQL-v0.1-Qwen-32B\footnote{\url{https://huggingface.co/onekq-ai/OneSQL-v0.1-Qwen-32B}} & 64.60 & 63.33 \\
        SuperSQL~\cite{li2024dawn} & 58.50 & 62.66 \\
        SFT CodeS-15B~\cite{li2024codes} & 58.47 & 60.37 \\
    \end{tabular}
    \vspace*{0.2cm}
    \caption{Results on the BIRD benchmark in the single-model, single-call category (i.e., no self-consistency). Accuracy is reported as exact-match execution accuracy on the dev and test sets. A dash (``--'') indicates results not computed. Our full method (``Databricks-RLVR-32B'') achieves the best test accuracy, while TAO alone (``Qwen2.5-32B-Coder-Instruct + TAO'') already reaches a competitive 67.4\%—roughly top-10 at the time of writing.}
    \label{tab:bird-results-single-call}
\end{table}

Our model achieves the best test set performance in this single-call setting. Accuracy follows BIRD’s exact-match execution metric.%
Although our model also had the highest dev set performance, it generalizes well to the test set (70.80\% vs.~73.56\%), showing stronger generalization than other top submissions. 

We also noticed that performing limited TAO training already gets us a non-trivial improvement of 67.40\%, which at the time of writing would have placed it in the top 10 on the leaderboard by dev set performance. We only used limited TAO training since our preliminary experiments showed that over-training led to less overall gains in the later online RL stage.

\subsection{Results with test-time compute (self-consistency)}
\pref{tab:bird-results-multi-call} shows the results with self-consistency. 
We generate 7 responses from Databricks-RLVR-32B and use a parameter-free weighted majority vote based approach to pick the best response. %
We achieve state-of-the-art results in the BIRD single-model category by using self-consistency on top of our Databricks-RLVR-32B model. In fact we do so while requiring fewer LLM calls (7) than the next best submission (8–32).\looseness=-1 %

\begin{table}[!htb]
    \centering
    \renewcommand{\arraystretch}{1.3}
    \setlength{\tabcolsep}{10pt}
    \begin{tabular}{l|cc|c}
        & \multicolumn{2}{c|}{\textbf{Accuracy (exact-match)}} & \textbf{Num.} \\
        \textbf{Model} & \textbf{Dev set} & \textbf{Test set} & \textbf{responses} \\
        \hline\hline
        \textbf{Databricks-RLVR-32B} & \textbf{--} & \textbf{75.68} & \textbf{7} \\
        Sophon-Text2SQL-32B & 72.43 & 74.79 & 8--32 \\
        Arctic-Text2SQL-R1-32B~\cite{yao2025arctic} & 72.20 & 73.84 & 1--7 \\
        Arctic-Text2SQL-R1-14B~\cite{yao2025arctic} & 71.40 & 72.22 & 1--7 \\
        SIFT-32B & 70.08 & 70.93 & $>$32 \\
        CrazyData-Text2SQL-32B & - & 70.71 & 8--32\\
        Infly-RL-SQL-32B & 70.08 & 70.60 & 1--7 \\
        Arctic-Text2SQL-R1-7B~\cite{yao2025arctic} & 70.70 & 70.43 & 1--7 \\
        XiYanSQL-QwenCoder-32B~\cite{liu2025xiyan} & 67.01 & 69.03 & 1--7 \\
    \end{tabular}
    \vspace*{0.2cm}
    \caption{Results on the BIRD benchmark in the single-model category with self-consistency. Accuracy is reported as exact-match execution accuracy on the dev and test sets; ``--'' indicates results not computed. The ``Num.~responses`` column denotes the number of LLM generations used for self-consistency. The BIRD leaderboard bins this value into three categories: ``few'' (1--7), ``many'' (8--32), and ``scale'' ($>$ 32). Our method (``Databricks-RLVR-32B'') achieves the best test accuracy (75.68\%) while requiring only 7 responses, compared to 8--32 for the next best system.}  
    \label{tab:bird-results-multi-call}
\end{table}

Finally, it's worth stressing that benefits of RLVR and self-consistency are complimentary as demonstrated.

We provide sample generations in the Appendix.

\section{Related work}
\label{sec:related_work}

\paragraph{Reinforcement learning and reasoning models.}
Recent work has increasingly explored using reinforcement learning to train models that perform multi-step reasoning, rather than relying solely on test-time computation \citep{snell2025scaling}. Common approaches use algorithms such as PPO \citep{schulman2017ppo} or GRPO \citep{shao2024deepseekmath}. Reinforcement learning was popularized in domains like games \citep{mnih2015dqn, silver2016mastering, silver2017mastering,vinyals2019grandmaster}, protein folding \citep{senior2020alphafold, jumper2021alphafold}, and code generation \citep{chen2021codex, le2022coderl, li2022alphacode, li2023alphacode2, dou2024stepcoder}. In the LLM setting, Reinforcement Learning from Human Feedback (RLHF) typically uses a reward model trained from human preference data, whereas RLVR replaces human preferences with verifiable signals from the environment---such as whether generated SQL produces the correct answer. %
Prior work in this area \citep{ma2025sqlr1, yao2025arctic} has applied RL to text-to-SQL and related code generation tasks. In contrast, we use a straightforward reward definition and combine offline and online RL strategies to fine-tune a single model, achieving state-of-the-art performance on BIRD.

\paragraph{Text-to-SQL.}
Early text-to-SQL systems were largely rule-based, relying on hand-written grammars and schema-specific heuristics. The introduction of large language models has moved the field toward training models that can \textit{directly} translate natural language questions into SQL queries, \textit{without} relying on manual rules. Many recent systems follow an \textit{agentic design}, combining a (fine-tuned) base model with components for schema linking (e.g., value retrieval), self-reflection, orchestration, and tightly engineered prompts (see, e.g., \cite{liu2025xiyan, pourreza2025chasesql}). %
Many prior works also put considerable effort into carefully curating large fine-tuning datasets; in contrast, we use a relatively small and simple dataset, relying on reinforcement learning to do most of the heavy lifting. Our main contribution is to show that reinforcement learning, even with this minimal setup, can deliver state-of-the-art results for text-to-SQL. A few recent papers have taken a similar reinforcement learning-based direction for text-to-SQL \citep{ma2025sqlr1, yao2025arctic}, and our work builds on this trend.

\paragraph{Real-world enterprise applications of text-to-SQL.}
Natural language interfaces to structured data have been deployed in a variety of settings. Early examples include Microsoft Power BI’s Q\&A and ThoughtSpot, which relied on symbolic techniques such as pattern matching and rule-based parsing. More recently, LLM-based systems have been integrated into general-purpose copilots (e.g., GitHub Copilot \citep{github2021copilot}, OpenAI’s Code Interpreter \citep{openai2023plugins} and reasoning models such as o1/o3 \citep{oai2024o1, oai2025o3o4mini}) as well as tools for data analysis (e.g., Databricks' Genie \citep{databricks2025genie} and Snowflake’s Cortex Analyst \citep{snowflake2024cortexanalyst}). The range of deployed systems is too broad to cover in full, but these examples illustrate the growing utility of text-to-SQL in both developer and analyst workflows. Our results on BIRD suggest that even a single-model system, trained with a simple reinforcement learning setup, can reach state-of-the-art performance, making this approach a promising candidate for practical deployment.

\section{Conclusion}
\label{sec:discussion}
In this report, we showed the power of Databricks RLVR and TAO fine-tuning services. We show that our approach when applied to BIRD -- a popular data science benchmark led to a new state-of-the-art performance with our first submission. The approach here is very general-purpose and can be applied to many other enterprise domains including in data science, coding, and business intelligence tasks. Our approach will be rolling in our new Agent Bricks product~\cite{agentbrick2025} and will be available to Databricks customers.

\bibliographystyle{plain}
\bibliography{main}

\newpage

\section{Appendix}

\subsection{Contributing authors}
The following members of the Databricks Mosaic Research and Genie teams contributed to this work (listed alphabetically):\\

Alnur Ali, Ashutosh Baheti, Jonathan Chang, Ta-Chung Chi, Brandon Cui, Andrew Drozdov, Jonathan Frankle, Abhay Gupta, Pallavi Koppol, Sean Kulinski, Jonathan Li, Dipendra Misra, Krista Opsahl-Ong, Jose Javier Gonzalez Ortiz, Matei Zaharia, and Yue Zhang.

\subsection{Sample model generations}
We list some successful examples below. The first successful example is fairly easy. The second one performs a table join and also shows model correcting a generation.

\begin{tcolorbox}[
width=\linewidth,
colback=succ!5!white, 
breakable,
colframe=green!75!black, 
title=Example 1 (Success)
]  
\textbf{Question:} \emph{``What is the highest eligible free rate for K-12 students in the schools in Alameda County?"}\\
\textbf{Evidence:} \emph{``Eligible free rate for K-12 = `Free Meal Count (K-12)` / `Enrollment (K-12)`"}\\
\textbf{Database:} california\_schools\\

\textbf{Response:} To determine the highest eligible free rate for K-12 students in the schools in Alameda County, we need to follow these steps:\\ \\1. Identify the relevant table and columns. In this case, the `frpm` table contains the necessary information about the free meal counts and enrollments for K-12 students.\\2. Filter the data to include only schools in Alameda County.\\3. Calculate the eligible free rate for each school using the formula: `Free Meal Count (K-12)` / `Enrollment (K-12)`.\\4. Find the maximum value of the calculated eligible free rate.\\\\Let\'s translate these steps into an SQL query:\\\\
\textasciigrave\textasciigrave\textasciigrave sql
\\SELECT MAX("Free Meal Count (K-12)" / "Enrollment (K-12)") AS MaxEligibleFreeRate\\FROM frpm\\WHERE "County Name" = \'Alameda\';\\
\textasciigrave\textasciigrave\textasciigrave
\\\\This query will return the highest eligible free rate for K-12 students in the schools in Alameda County.
\end{tcolorbox}

\begin{tcolorbox}[
width=\linewidth,
colback=succ!5!white, 
breakable,
colframe=green!75!black, 
title=Example 2 (Success)
]  
\textbf{Question:} \emph{``Among the superheroes with the race of god/eternal, how many of them are male"}\\
\textbf{Evidence:} \emph{``race "god/eternal" refers to race\_id = 21; male refers to gender.id = 1"}\\
\textbf{Database:} superhero\\

\textbf{Response:} To solve this problem, we need to follow these steps:\\

1. Identify the table that contains the superhero information, including their race and gender. In this case, it's the `superhero` table.\\
2. Determine the tables that contain the race and gender information. Here, the `race` table contains the race information, and the `gender` table contains the gender information.\\
3. Join the `superhero` table with the `race` table on the `race\_id` to filter superheroes with the race "god/eternal".\\
4. Join the `superhero` table with the `gender` table on the `gender\_id` to filter superheroes who are male.\\
5. Count the number of superheroes that meet both conditions (race "god/eternal" and gender male).\\

Let's translate these steps into an SQL query:\\

\textasciigrave\textasciigrave\textasciigrave sql\\
SELECT COUNT(*)\\
FROM superhero AS h\\
JOIN race AS r ON h.race\_id = r.id\\
JOIN gender AS g ON h.gender\_id = g.id\\
WHERE r.race = ``God/Eternal" AND g.gender = ``Male";\\
\textasciigrave\textasciigrave\textasciigrave\\

However, the question specifies that "race 'god/eternal'" refers to `race\_id = 21`. We can use this information directly in the WHERE clause to avoid joining the `race` table. Here is the optimized query:\\

\textasciigrave\textasciigrave\textasciigrave sql\\
SELECT COUNT(*)\\
FROM superhero AS h\\
JOIN gender AS g ON h.gender\_id = g.id\\
WHERE h.race\_id = 21 AND g.gender = 'Male';\\
\textasciigrave\textasciigrave\textasciigrave\\

This query will give us the count of male superheroes with the race "god/eternal".
\end{tcolorbox}

We also list an unsuccessful example below. In this case, the model makes two mistakes. It uses type of foreign$\_$data table instead of the card's table which has different value even though the names are confusing. Second, it forgot to apply distinct to the returned values.

\begin{tcolorbox}[
width=\linewidth,
colback=abstract!5!white, 
colframe=red!75!black, 
breakable,
title=Example 3 (Failure)
]  
\textbf{Question:} \emph{``What's the German type of the card `Ancestor's Chosen'?"}\\
\textbf{Evidence:} \emph{``German refers to language = 'German'; "Ancestor's Chosen" refers to name = 'Ancestor's Chosen'"}\\
\textbf{Database:} card\_games\\

\textbf{Response:} To determine the German type of the card "Ancestor's Chosen", we need to join the `cards` table with the `foreign\_data` table. The `cards` table contains the English name of the card, and the `foreign\_data` table contains the translated types in different languages. We will filter the results to only include the card with the name "Ancestor's Chosen" and the language set to 'German'.\\

Here are the steps to construct the SQL query:\\

1. Identify the table that contains the card names and their corresponding UUIDs (`cards`).\\
2. Identify the table that contains the translated types and their corresponding UUIDs (`foreign\_data`).\\
3. Join the `cards` table with the `foreign\_data` table on the UUID column.\\
4. Filter the joined table to only include rows where the card name is "Ancestor's Chosen" and the language is 'German'.\\
5. Select the `type` column from the `foreign\_data` table.\\

Let's translate these steps into an SQL query:\\

\textasciigrave\textasciigrave\textasciigrave sql\\
SELECT fd.type\\
FROM cards c\\
JOIN foreign\_data fd ON c.uuid = fd.uuid\\
WHERE c.name = ``Ancestor's Chosen" AND fd.language = ``German";\\
\textasciigrave\textasciigrave\textasciigrave\\

\end{tcolorbox}

\end{document}